\documentclass[times]{article}
\author{Christophe Lohou}

\usepackage{arxiv}

\usepackage[utf8]{inputenc} 
\usepackage[T1]{fontenc}    
\usepackage[hidelinks]{hyperref}       
\usepackage{url}            
\usepackage{booktabs}       
\usepackage{amsfonts}       
\usepackage{nicefrac}       
\usepackage{microtype}      
\usepackage{cleveref}       
\usepackage{graphicx}
\usepackage{doi}
\usepackage{longtable}
\usepackage{supertabular}
\usepackage[english]{babel}
\usepackage[table,xcdraw]{xcolor}
\usepackage{breakurl}

\usepackage{amssymb}
\usepackage{latexsym}
\usepackage{psfrag} 
\usepackage{epsfig}

\graphicspath{ {./} }
\begin{document}

\title{Topological Numbers \\and their use to characterize simple points\\ 
for $2$D binary images}
\author{
\href{https://orcid.org/0000-0001-5352-8237}{
    \hspace{1mm}Christophe Lohou}\\
    Université Clermont Auvergne,\\
    Clermont Auvergne INP,\\
    CNRS, Institut Pascal\\
    F-63000 Clermont-Ferrand, France\\
	\texttt{christophe.lohou@uca.fr}
}

\renewcommand{\headeright}{}
\renewcommand{\undertitle}{}
\renewcommand{\shorttitle}{}

\date{\today}
\maketitle
\begin{abstract}

In this paper, we adapt the two topological numbers, which have been proposed to efficiently characterize simple points in specific neighborhoods for $3$D binary images,  
to the case of $2$D binary images. 

Unlike the $3$D case, we only use a single neighborhood to define these two topological numbers for the $2$D case.
Then, we characterize simple points either by using the two topological numbers 
or by a single topological number linked to another one condition. 

We compare the characterization of simple points by topological numbers 
with two other ones based on Hilditch crossing number and Yokoi number.

We also highlight the number of possible configurations corresponding to a simple point, which also represents the maximum limit  
of local configurations that a thinning algorithm operating by parallel deletion of simple (individual) points may delete while preserving topology
(limit usually not reachable, depending on the deletion strategy).

\end{abstract}


\section{Introduction}
 
Our current research on segmentation in medical imaging 
lead us to propose different operators for processing $2$D/$3$D images, or $2$D+t/$3$D+t sequences,
for the binary or grayscale cases, operators that must respect certain topological constraints.
For example, one of our operators must ensure that it well preserves topology of a $2$D binary image 
while simultaneously deleting several points of that image. 

Such similar operators have been proposed in Digital Topology framework \cite{KoRo89}.
This is for example the case with skeletonization algorithms (e.g. \cite{LoBe2005}), 
based on the removal of simple points,
points whose sequential removal preserves the topology of the image.

In 1995, Bertrand proposed the notion of topological numbers \cite{Be94}: 
these allow to count the number of related components of the object and its complementary
in several specific neighborhoods of a given point.
A characterization of simple points using these topological numbers 
was also proposed for $3$D simple points \cite{BeMa94}.

In this article, we recall the basic notions of Digital Topology (Sect. \ref{sec:basic_notions}),
then we define the notion of topology preservation and the one of of a simple point for the $2$D case (Sect. \ref{sec:topology_preservation}).
In Sect. \ref{sec:topological_numbers}, we propose the adaptation of the topological numbers to the $2$D case, 
by only using a single neighborhood, in contrary to the $3$D case.
Then, in Sect. \ref{sec:localcharacterization}, we propose the characterization of simple points for $2$D binary images by using these two topological numbers; 
in Sect.\ref{sec:efficient_characterization}, we propose a characterization of simple points by the computing of a single topological number and another one condition. 
In Sect. \ref{sec:comparisons}, we compare our approach with two another propositions of characterizations of simple points. 
Eventually, we conclude in Sect. \ref{sec:conclusion}, with the notion of \textit{rate of deletability} that an algorithm, operating by parallel deletion of individual simple points, may reach, 
and with the notion of \textit{more powerful thinning algorithm}.

\section{Basic notions}\label{sec:basic_notions}

\begin{figure}
\psfrag{c}{$a$}
\psfrag{d}{$b$}
\psfrag{e}{$c$}
\psfrag{f}{$d$}
\psfrag{g}{$e$}
\psfrag{h}{$f$}
\psfrag{i}{$g$}
\psfrag{a}{(a)}
\psfrag{b}{(b)}
\begin{center}
\includegraphics[width=8cm]{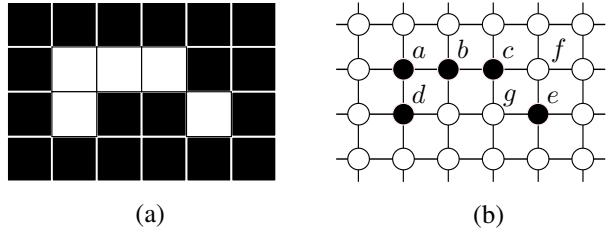}
\end{center}
\caption{(a) A $2$D binary image, (b) a corresponding mapping to $\mathcal{Z}^2$.}
\label{fig:image_2D}
\end{figure}

A $2$D binary image consists of square elements, \textit{pixels},
which can be black for the background and white for the forms of interest, 
also called the \textit{object} afterwards (Fig. \ref{fig:image_2D} (a)). 
A pixel can be mapped to a point in the grid with integer coordinates, 
$\mathcal{Z}^2$, Cartesian product of $\mathcal{Z}$ by $\mathcal{Z}$. 
By convention, we associate in $\mathcal{Z}^2$ 
a black point for a white pixel of the image - thus, to an element of the object -,
and a white point for a black pixel of the image (Fig. \ref{fig:image_2D} (b)).

\begin{figure}
\psfrag{x}{$x$}
\psfrag{a}{(a)}
\psfrag{b}{(b)}
\psfrag{c}{(c)}
\psfrag{d}{(d)}
\begin{center}
\includegraphics[width=8cm]{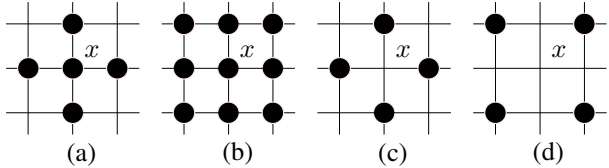}
\end{center}
\caption{(a) $N_4(x)$, (b) $N_8(x)$, (c) $4$-neighbors of $x$, (d) $8$-neighbors of $x$.}
\label{fig:voisinage_2D}
\end{figure}

Let $x(x_1,x_2)$ be the point of coordinates $(x_1,x_2)$ in $\mathcal{Z}^2$. 
Let $x(x_1,x_2)$ and $y (y_1,y_2)$ be two points of $\mathcal{Z}^2$. 
The following two distances can be defined: 
$d_4(x,y)=\sum_{i=1}^{2}|y_i-x_i|$, 
and $d_8(x,y)=\max_{i=1,2} |y_i-x_i|$. 
We define the two following neighborhoods: 
let $x \in \mathcal{Z}^2$, 
$N_4(x)=\{y \in \mathcal{Z}^2, d_4(x,y)\leq 1 \}$ (Fig. \ref{fig:voisinage_2D} (a)), 
and $N_8(x)=\{y \in \mathcal{Z}^2, d_8(x,y)\leq 1 \}$ (Fig. \ref{fig:voisinage_2D} (b)). 
Note $N_n^*(x)=N_n(x) \setminus \{x\}$, for $n \in  \{4,8\}$.  
The $n$ points of $N_n^*(x)$ define the $n$-neighborhood. 
Two points $x$ and $y$ of $\mathcal{Z}^2$ are \textit{$n$-adjacent} if $y \in N_n^*(x)$, $n \in \{4,8\}$. 
The $4$ points of $N_4^*(x)$ are the \textit{$4$-neighbors of $x$} (Fig. \ref{fig:voisinage_2D} (c)), 
the $4$ points of $N_8^*(x) \setminus N_4^*(x)$ are the \textit{$8$-neighbors of $x$} (Fig. \ref{fig:voisinage_2D} (d)).

Let us recall Jordan's theorem in $\mathcal{R}^2$: 
any simple closed curve separates the plane into two domains which are the interior domain and the exterior domain of the curve.
In order to verify it in the discrete case, we need to use the $4$-adjacency for black points 
and the $8$-adjacency for white points or vice versa \cite{KoRo89}. 
We can then define a \textit{digital image} as the data ($\mathcal{Z}^2, n, \overline{n}, X$),
with $X \subseteq \mathcal{Z}^2$ as the object, 
$(n,\overline{n})=(4,8)$ or $(8,4)$ \cite{KoRo89}; 
the points of $\mathcal{Z}^2$ are the points of the image, 
the points of $X$ are the black points of the image,
and the points of $\overline{X}=\mathcal{Z}^2 \setminus X$, the \textit{complementary of $X$} in the image, 
are the white points of the image.

Two black points are \textit{adjacent} if they are $n$-adjacent, 
two white points or one white point and a black point are \textit{adjacent} 
if they are $\overline{n}$-adjacent. A \textit{$n$-path} is a sequence 
$<p_i; 0 \leq i \leq l>$ 
of points such that  
$p_i$ is $n$-adjacent to $p_{i+1}$ for any $0 \leq i < l$, 
if $p_0=p_l$, the path is  said to be \textit{closed}. 
Let $X \subseteq \mathcal{Z}^2$, 
two points $p$ and $q$ of $X$ 
are \textit{$n$-connected into $X$} 
if and only if there is a $n$-path included in $X$ 
which links $p$ to $q$. 
The relation "to be $n$-connected in $X$" is an equivalence relation,
the equivalence classes of this relation are the \textit{$n$-connected components} of the image. 
A $n$-connected component of the set of black points of the image is called a \textit{black component}
and a $\overline{n}$-connected component of the set of white points is called a \textit{white component}.
In a finite digital image (when $X$ is a finite set),
there is a single infinite white component called \textit{background} of the image.
The finite white components are called \textit{holes}. 
A black point is said to be \textit{$n$-isolated} if it is not $n$-adjacent to any other black point. 
A black point is said to be a \textit{$n$-interior} point if it is not $\overline{n}$-adjacent to any white point.
 
Consider Figure \ref{fig:image_2D} (b). 
If $(n, \overline{n})=(4,8)$, points $c$ and $e$ are not $4$-adjacent, 
there are two $4$-connected components of $X$ ($\{a,b,c,d\},\{e\}$), 
note that points $f$ and $g$ of $\overline{X}$ are $8$-adjacent 
and separate these two $4$-connected components 
(the analog of Jordan's theorem in the $2$D discrete case is verified). 
If $(n,\overline{n})=(8,4)$, 
there is only a single $8$-connected component of $X$ ($\{a,b,c,d,e\}$), 
note that the points $f$ and $g$ of $\overline{X}$ are not $4$-adjacent and 
cannot "cut" the connection between the points $c$ and $e$ of $X$ 
(the analog of Jordan's theorem in the $2$D discrete case is verified). 

\section{Topology preservation, simple points}\label{sec:topology_preservation}

Let $(\mathcal{Z}^2, n, \overline{n}, X)$, with $X \subseteq \mathcal{Z}^2$  be a digital image. 
A point $x \in X$ is said to be \textit{$n$-simple for $X$} 
if its deletion from $X$ 
\textit{well preserves the topology of the image}  
in the sense that both 
the number of $n$-connected components of the object $X$ 
and the number of $\overline{n}$-connected components of its complementary $\overline{X}$ 
are the same before and after the deletion of the point $x$ \cite{Mor81,KoRo89}. 

\begin{figure}[!t]
\psfrag{x}{$x$}
\psfrag{a}{$a$}
\psfrag{b}{$b$}
\psfrag{c}{$c$}
\psfrag{d}{$d$}
\psfrag{e}{$e$}
\psfrag{f}{$f$}
\psfrag{g}{$g$}
\psfrag{h}{$h$}
\psfrag{i}{(a)}
\psfrag{j}{(b)}
\psfrag{k}{(c)}
\begin{center}
\includegraphics[width=8cm]{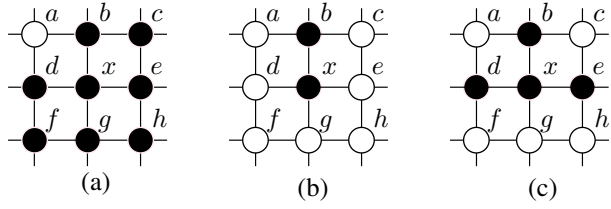}
\end{center}
\caption{(a) $x$ is $4$-simple for $X$ and is not $8$-simple for $X$, 
(b) $x$ is both $4$-simple for $X$ and $8$-simple for $X$, 
(c) $x$ is $8$-simple for $X$ and is not $4$-simple for $X$.}
\label{fig:exemples_simples_ou_non_2D}
\end{figure}

Let us consider Fig. \ref{fig:image_2D} (b). 
For $n=4$, 
the image contains two $4$-connected components of $X$ ($\{a,b,c,d\}$ and $\{e\}$),
and a single $8$-connected component of $\overline{X}$. 
After the deletion of the point $a$ from the object $X$, 
the resulting image will contain three $4$-connected components of $X \setminus \{a\}$ 
($\{b,c\}$, $\{d\}$ ,$\{e\}$), thus the topology would not be preserved, therefore the point $a$ is not $4$-simple for $X$.
For $n=8$, 
the image contains a single $8$-connected component of $X$ ($\{a,b,c,d,e\}$),
and a single $4$-connected component of $\overline{X}$. 
After the deletion of $a$ from the object $X$, 
the resulting image will still contain both one $8$-connected component of $X \setminus \{a\}$ 
($\{b,c, d, e\}$), and one $4$-connected component of $\overline{X} \cup \{a\}$. 
Thus the topology would be preserved, therefore the point $a$ is $8$-simple for $X$.

Let us consider other images, those in Fig. \ref{fig:exemples_simples_ou_non_2D}, that we will also use later:  
\begin{itemize}
\item (a). 
For $n=4$, 
the image contains a single $4$-connected component of $X$ ($\{b,c,d,x,e,f,g,h\}$)
and a single $8$-connected component of $\overline{X}$ ($\{a\}$). 
After the deletion of $x$ from the object $X$, 
the resulting image will contain a single $4$-connected component of $X \setminus \{x\}$ 
($\{b,c,d,e,f,g,h\}$)
and a single $8$-connected component of $\overline{X} \cup \{x\}$ ($\{a,x\}$). 
Thus, both the numbers of $4$-connected components of $X$ and of $X \setminus \{x\}$ 
are unchanged,  
and the numbers of $8$-connected components of $\overline{X}$ 
and of $\overline{X} \cup \{x\}$ are unchanged, 
therefore $x$ is $4$-simple for $X$.   

For $n=8$, 
the image contains a single $8$-connected component of $X$ ($\{b,c,d,x,e,f,g,h\}$)
and a single $4$-connected component of $\overline{X}$ ($\{a\}$). 
After the deletion of $x$ from $X$, 
the resulting image will contain two $4$-connected components of $\overline{X} \cup \{x\}$ 
($\{a\}$ and $\{x\}$, the point $x$ not being $4$-adjacent to the point $a$). 
Since the numbers of white components are not the same before and after the deletion of $x$, 
then the topology would not be preserved,  
therefore the point $x$ is not $8$-simple for $X$. 

\item (b). 
The image contains a single $n$-connected component of $X$ 
and a single $\overline{n}$-connected component of $\overline{X}$, 
the removal of $x$ from $X$ does not change these two numbers; 
thus the point $x$ is $n$-simple for $X$, for $n \in \{4,8\}$. 

\item (c). 
The image contains a single $n$-connected component of $X$ ($\{b,d,x,e\}$) 
and three $\overline{n}$-connected components of $\overline{X}$ 
($\{a\}$, $\{c\}$, $\{f, g, h\}$), for $n \in \{4,8\}$.

For $n=4$, the deletion of $x$ from $X$ 
will break the component of $X$ into three $4$-connected components of $X \setminus \{x\}$ 
($\{b\}$, $\{d\}$, $\{e\}$), 
and will merge the three $8$-connected components of $\overline{X}$ 
into a single $8$-connected component of $\overline{X} \cup  \{x\}$ 
($\{a, c, x, f, g, h\}$). 
Thus the topology would not be preserved, 
therefore the point $x$ is not $4$-simple for $X$.  

For $n=8$, after the deletion of $x$ from $X$, 
it remains a single $8$-connected component of $X \setminus \{x\}$ ($\{b,d,e\})$ 
and three $4$-connected components of $\overline{X} \cup \{x\}$ 
($\{a\}$, $\{c\}$, $\{x, f, g, h\}$, 
the point $x$ being $4$-adjacent to the point $g$). 
Thus, both the numbers of $8$-connected components of $X$ and of $X \setminus \{x\}$ 
are the same,  
the numbers of $4$-connected components of $\overline{X}$ and of $\overline{X} \cup \{x\}$ are the same, 
therefore $x$ is $8$-simple for $X$. 
\end{itemize}

\section{Topological numbers}\label{sec:topological_numbers}

In the previous section, 
in order to check whether a point is simple or not, 
we counted the number of components of the object and its complementary in the (overall) image before and after the removal of that point. 
In fact, we have the remarkable property of being able to locally verify 
whether a point is simple or not 
by the only examination of $N^*_8(x)$ \cite{KoRo89}. 

Bertrand has introduced the topological numbers for $3$D binary images \cite{Be94}: 
topological numbers are the number of connected components of the object and its complementary 
in several specific neighborhoods of a point 
and allow to efficiently characterize whether a point $x$ is simple or not \cite{BeMa94}. 
In this section, we give the $2$D version of these numbers by only using a single neighborhood ($N^*_8(x)$ of a point $x$), 
and we will show in the next section, that these definitions of topological numbers may allow us 
to characterize simple points for $2$D binary images.

Let $X \subset \mathcal{Z}^2$ and $x \in X$. 
Note $C_n(X)$ the number of $n$-connected components of $X$ 
and $C_n^x(X)$ the number of $n$-connected component of $X$ 
and $n$-adjacent to the point $x$, 
the cardinal number of $X$ is denoted by $\#X$. 

\textbf{Definition$1$ (topological numbers in $2$D):} 
Let $X \subset \mathcal{Z}^2$ and $x \in X$. 
The \textit{topological numbers of $X$ and $x$} 
are the two numbers: 
$T_n(x,X)=\#C_n^x[X \cap N^*_8(x)]$, $n \in \{4,8\}$. 

More precisely, 
the topological number for $n=8$ 
is the number $T_8(x,X)=\#C_8^x[X \cap N^*_8(x)]=\#C_8[X \cap N^*_8(x)]$ 
(since a $8$-connected component included in $N^*_8(x)$ is necessarily $8$-adjacent to $x$); 
the topological number for $n=4$ 
is the number $T_4(x,X)=\#C_4^x[X \cap N^*_8(x)]$. 

\begin{table}
\centering
\renewcommand{\arraystretch}{1.25}
\begin{tabular}{|l||c|c|c|c|c|c|}
\hline
      & $k=0$ & $k=1$ & $k=2$ & $k=3$ & $k=4$ & $k>4$\\ 
\hline
$T_4(x,X)$ & $16$ & $117$ & $102$ & $20$ & $1$  & $0$\\
\hline
$T_8(x,X)$ & $1$ & $132$ & $102$ & $20$  & $1$ & $0$\\
\hline
\hline
      & $k=0$ & $k=1$ & $k=2$ & $k=3$ & $k=4$ & $k>4$\\ 
\hline
$T_8(x,\overline{X})$ & $1$ & $132$ & $102$ & $20$  & $1$ & $0$\\
\hline
$T_4(x,\overline{X})$ & $16$ & $117$ & $102$ & $20$ & $1$  & $0$\\
\hline
\end{tabular}
\caption{Number of configurations in $N^*_8(x)$ of a point $x$ which belongs to $X$, 
such that $T_n(x,X)=k$ or $T_{\overline{n}}(x,\overline{X})=k$ with $k \in \mathbb{N}$, according to $n \in \{4, 8\}$. 
}
\label{tab:tab_nombre_topo_T_ou_Tbarre}
\end{table}

In the following, we call \textit{(local) configuration} of a point $x$, the neighborhood $N^*_8(x)$ of a point $x$.

By reviewing with a computer 
all $2^8=256$ configurations of black and white points 
in the neighborhood $N^*_8(x)$ of a point $x$, 
we may compute the number of configurations having specific values of $T_n(x,X)$ or $T_{\overline{n}}(x,\overline{X})$
for $n \in \{4, 8\}$, see Tab. \ref{tab:tab_nombre_topo_T_ou_Tbarre}.

Let us consider an image $\mathcal{I}_1=(\mathcal{Z}^2, n, \overline{n}, X)$, 
with $X \subset \mathcal{Z}^2$ and $x \in X$, 
and the image $\mathcal{I}_2=(\mathcal{Z}^2, m, \overline{m}, Y=\mathcal{Z}^2 \setminus X)$, with $Y \subset \mathcal{Z}^2$ obtained by changing black points of $\mathcal{I}_1$ into white points, except $x$, and white points into black points.
Let us assume that $T_n(x,X)=k_1$, with $k_1 \in \mathbb{N}$, it means that there is $k_1$ $n$-connected components of $X$ (black points) and $n$-adjacent to $x$, 
thus there is $k_1$ $n$-connected components of $\overline{Y}$ (white points) and $n$-adjacent to $x$, 
therefore $T_{\overline{m}}(x,\overline{Y})=k_1$, with $\overline{m}=n$.
In the same way, let us assume that $T_{\overline{n}}(x,\overline{X})=k_2$, with $k_2 \in \mathbb{N}$, it means that there is $k_2$ $\overline{n}$-connected components of $\overline{X}$ (white points) 
and $\overline{n}$-adjacent to $x$, 
thus there is $k_2$ $\overline{n}$-connected components of $Y$ (black points) and $\overline{n}$-adjacent to $x$, 
therefore $T_{m}(x,Y)=k_2$, with $m=\overline{n}$. Thus, we have the following proposition.

\textbf{Proposition$1$:}   
The number of configurations such that 
$T_n(x,X)=k$ is the same as the number of configurations such that 
$T_{\overline{n}}(x,\overline{X})=k$, with $k \in \mathbb{N}$, with $n \in \{4, 8\}$.

In fact, we may deduce the bottom of Tab. \ref{tab:tab_nombre_topo_T_ou_Tbarre} (the number of configurations such that $T_{\overline{n}}(x,\overline{X})=k$ 
from the top of this same table (the number of configurations such that $T_n(x,X)=k$) with Proposition$1$.

\section{Local characterization of simple points by the computing of topological numbers}\label{sec:localcharacterization}

In $3$D, we may characterize a simple point if and only if its two topological numbers are equal to $1$.

By reviewing with a computer 
all $256$ configurations of black and white points 
in the neighborhood $N^*_8(x)$ of a point $x$, 
we have verified that any configuration such that the topology is well preserved 
before and after the deletion of $x$ 
(by verifying the preservation of numbers of connected components 
of an object $X$ and its complement $\overline{X}$ 
before and after 
the deletion of the point $x$) 
if and only if both $T_n(x,X)=1$ and $T_{\overline{n}}(x,\overline{X})=1$, 
for $n \in \{4, 8\}$. 
We highlight that for the $2$D case, only one neighborhood ($N^*_8(x)$) is sufficient in contrary to the $3$D case.

\textbf{Proposition$2$ (local characterization of a $4$-simple point):}  
Let $X \subseteq \mathcal{Z}^2$ and $x \in X$. 
The point $x$ is a $4$-simple point for $X$ $\Leftrightarrow$ $T_4(x,X)=1$ and $T_8(x,\overline{X})=1$. 

\textbf{Proposition$3$ (local characterization of a $8$-simple point):}  
Let $X \subseteq \mathcal{Z}^2$ and $x \in X$. 
The point $x$ is a $8$-simple point for $X$ $\Leftrightarrow$ $T_8(x,X)=1$ and $T_4(x,\overline{X})=1$. 

Therefore, we have the following Proposition:

\textbf{Proposition$4$ (local characterization of a $n$-simple point):}  
Let $X \subseteq \mathcal{Z}^2$ and $x \in X$. 
The point $x$ is a $n$-simple point for $X$ $\Leftrightarrow$ $T_n(x,X)=1$ and $T_{\overline n}(x,\overline{X})=1$. 

Among the $256$ possible configurations,
$116$ configurations correspond to a $n$-simple point 
($45.31\%$ of possible configurations); 
thus $140$ configurations correspond to a non $n$-simple point 
($54.69\%$ of possible configurations).

Let us consider again the examples of Fig. \ref{fig:exemples_simples_ou_non_2D} 
to check the simplicity or not of points $x$ 
using topological numbers:
\begin{itemize}
\item (a). 
$N^*_8(x)\cap X$ contains a single $4$-connected component (thus $8$-connected too) of $X$ 
($\{b,c,d,e,f,g,h\}$), 
this component is $4$-adjacent to $x$, 
thus $T_4(x,X)=T_8(x,X)=1$.
$N^*_8(x) \cap \overline{X}$ contains only one $4$-connected component (thus $8$-connected too) 
of $\overline{X}$ ($\{a\})$, 
this component is not $4$-adjacent to $x$, 
however this component is $8$-adjacent to $x$, 
thus $T_4(x,\overline{X})=0$ and $T_8(x,\overline{X})=1$. 
Therefore, the point $x$ is $4$-simple for $X$, 
but is not $8$-simple for $X$. 
  
\item (b). 
$N^*_8(x) \cap X$ contains a single $4$-connected component 
(thus $8$-connected too) of $X$ ($\{b\}$)  
and this component is $4$-adjacent to the point $x$, thus $T_4(x,X)=T_8(x,X)=1$. 
$N^*_8(x) \cap \overline{X}$ contains a single $4$-connected component (thus $8$-connected too) 
of $\overline{X}$ ($\{a, c, d, e, f, g, h\}$) and it is $4$-adjacent to the point $x$;  
therefore $T_4(x,\overline{X})=T_8(x,\overline{X})=1$. 
The point $x$ is $4$-simple for $X$ and is $8$-simple for $X$ too. 

\item (c). 
For $n=4$, $N^*_8(x) \cap X$ is made of 
three $4$-connected components of $X$ that are $4$-adjacent to $x$ ($\{b\}, \{d\}, \{e\}$)
and $N^*_8(x) \cap \overline{X}$ contains three $8$-connected components of $\overline{X}$ 
that are $8$-adjacent to $x$ 
($\{a\}, \{c\}, \{f,g,h\}$)
thus $T_4(x,X)=3$ and $T_8(x,\overline{X})=3$, 
therefore the point $x$ is not $4$-simple for $X$. 
\\
For $n=8$, $N^*_8(x) \cap X$ is made of 
a single $8$-connected component of $X$ ($\{b,d,e\}$) 
and $N^*_8(x) \cap \overline{X}$ is made of three $4$-connected components of $\overline{X}$ 
($\{a\}, \{c\}, \{f,g,h\}$), 
but only the component $\{f,g,h\}$ is $4$-adjacent to $x$; 
thus $T_8(x,X)=1$ and $T_4(x,\overline{X})=1$, 
therefore the point $x$ is $8$-simple for $X$. 
\end{itemize}

\begin{figure}[!t]
\psfrag{x}{$x$}
\psfrag{a}{(a)}
\psfrag{b}{(b)}
\psfrag{c}{(c)}
\begin{center}
\includegraphics[width=7.5cm]{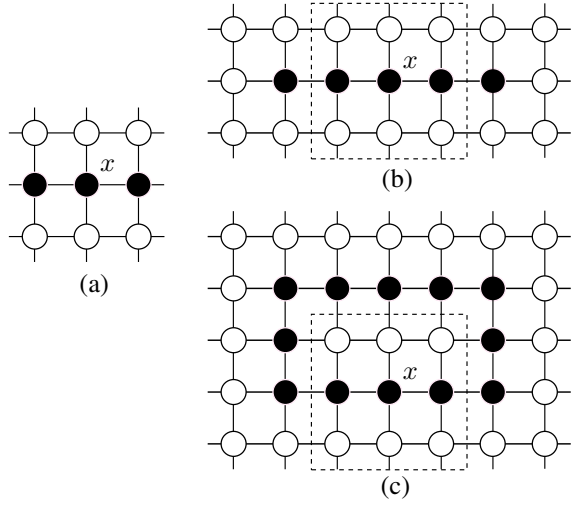}
\end{center}
\caption{(a) $x$ is not $n$-simple for $X$. 
It is not possible to locally (in $N^*_8(x)$) determine the reason why $x$ is not $n$-simple for $X$:  
(b) the deletion of $x$ will break the object $X$ onto two $n$-connected components, 
or (c) will merge two $\overline{n}$-connected components of $\overline{X}$.
}
\label{fig:raison_non_locale_2D}
\end{figure}

Although topological numbers allow us to efficiently 
determine whether a point $x$ of $X$ is simple or not 
for $X$ by only examining its local neighborhood $N^*_8(x)$,
note that it is possible not to know in $N^*_8(x)$ the reason why a point is not simple. 
Consider Figure \ref{fig:raison_non_locale_2D}.
In (a), we have $T_n(x,X)=2$ and
$T_{\overline{n}}(x,\overline{X})=2$,
for $n \in \{4,8\}$,
therefore the point $x$ is not a $n$-simple point for $X$. 
In (b), deleting $x$ will break the object $X$ onto two black components,
while in (c), deleting $x$ will merge two white components of $\overline{X}$, 
nethertheless we have the same local configuration of the neighborhood
$N^*_8(x)$ of the point $x$, described in (a),
for the two examples (b) and (c). 

\section{More efficient characterization of simple points with topological numbers}\label{sec:efficient_characterization}

We said that $116$ configurations correspond to a $n$-simple point $x$ for $X$, $n \in \{4, 8\}$, 
\textit{i.e.} $T_n(x,X)=1$ and $T_{\overline{n}}(x,\overline{X})=1$.
 
Let $n=4$. 
In Tab. \ref{tab:tab_nombre_topo_T_ou_Tbarre}, 
we may retrieve that a single configuration is such that $T_8(x,\overline{X}) = 0$.
This configuration is such that there is no white point $8$-adjacent to $x$; 
in other words, $x$ is a $4$-interior point for $X$. 
It corresponds to a single possibility that it is described in Fig. \ref{fig:voisinage_2D} (b).
That configuration is such that $T_4(x,X)=1$.
Then, we may also retrieve from Tab. \ref{tab:tab_nombre_topo_T_ou_Tbarre} that 
$117$ configurations are such that $T_4(x,X)=1$ .
Among these configurations, only one of them is such that $T_4(x,X)=1$ and $T_8(x,\overline{X}) \neq 1$.
This is precisely the configuration corresponding to a $4$-interior point. 

Thus we have the following proposition:

\textbf{Proposition$5$:} 
Let $X \subset \mathcal{Z}^2$ and $x \in X$, 
$x$ is $4$-simple for $X$ 
$\Leftrightarrow$ $T_4(x,X)=1$ and $x$ is not a $4$-interior point. 

Let $n=8$. 
In Tab. \ref{tab:tab_nombre_topo_T_ou_Tbarre}, 
we may retrieve that $16$ configurations are such that $T_4(x,\overline{X}) = 0$.
These configurations are such that there is no white point $4$-adjacent to $x$; 
in other words, $x$ is a $8$-interior point for $X$. 
It corresponds to $16$ possibilities, one of them is described in Fig. \ref{fig:exemples_simples_ou_non_2D} (a)
(there are $16$ possibilities to define the four $8$-neighbors according to their membership either to $X$ or to $\overline{X}$). 
These configurations are such that $T_8(x,X)=1$.
Then, we may also retrieve from Tab. \ref{tab:tab_nombre_topo_T_ou_Tbarre} that 
$132$ configurations are such that $T_8(x,X)=1$ (Tab. \ref{tab:tab_nombre_topo_T_ou_Tbarre}).
Among these configurations, only $16$ are such that $T_8(x,X)=1$ and $T_4(x,\overline{X}) \neq 1$.
These are precisely the configurations corresponding to a $8$-interior point. 
Thus we have the following property:

\textbf{Proposition$6$:} 
Let $X \subset \mathcal{Z}^2$ and $x \in X$,  
$x$ is $8$-simple for $X$ 
$\Leftrightarrow$ $T_8(x,X)=1$ and $x$ is not a $8$-interior point.

Therefore, we have the following proposition:

\textbf{Proposition$7$:} 
Let $X \subset \mathcal{Z}^2$ and $x \in X$, 
$x$ is $n$-simple for $X$ 
$\Leftrightarrow$ $T_n(x,X)=1$ and $x$ is not a $n$-interior point, $n \in \{4,8\}$.

We may consider that these three last characterizations (the computing of a single topological number and the check whether the point $x$ is a $n$-interior point or not) 
are more efficient than the computing of the two topological numbers as in Proposition$2$, Proposition$3$ and Proposition$4$.

\section{Comparisons}\label{sec:comparisons}

In this section, we recall both Hilditch crossing number and Yokoi number, 
we first compare them with topological number, 
then we compare characterization of simple points using these three numbers.

\subsection{Hilditch crossing number}

The Hilditch number $H(x,X)$, only proposed for $n=8$, 
is the sum of the number of passages from $0$ to $1$ 
when traveling through the local neighborhood of $x$ ($N^*_8(x)$), 
starting from a point and returning there, 
and such that both
the first point is a $4$-neighbor of $x$, 
and when there are two black points $4$-adjacent to $x$ and $8$-adjacent between them two, 
we directly pass from the first black point to the second one without passing through the $8$-neighbor (other than $x$) between them 
(in other words, we "cut the angles") \cite{KoRo89}.

In Tab.\ref{tab:tab_Hilditch}, are given the number of configurations 
corresponding to a point $x$ of $X$ such that 
 $H(x,X)=k$ with $k \in \mathbb{N}$;
 we also recall $T_n(x,X)=k$, with $k \in \mathbb{N}$, for $n=8$.

There are $116$ configurations such that $H(x,X)=1$, 
we have verified, with a computer, that these configurations precisely correspond to $8$-simple points for $X$.

\textbf{Proposition$8$:}
Let $X \subset \mathcal{Z}^2$ and $x \in X$, 
$x$ is $8$-simple for $X$ $\Leftrightarrow$ $H(x,X)=1$. 

We also have verified, with a computer, that any configuration in $N^*_8(x)$ is such that $H(x)=T_8(x,X)$ 
except for $16$ configurations: 
these are configurations corresponding to a $8$-interior point, 
in these cases $H(x,X)=0$ and $T_8(x,X)=1$.

Thus, we have the following Proposition:\\ 
\textbf{Proposition$9$:}
Let $X \subset \mathcal{Z}^2$ and $x \in X$. 
$H(x,X)=T_8(x,X)$ except for the $16$ configurations such that $x$ is a $8$-interior point, 
in this case $H(x,X)=0$ and $T_8(x,X)=1$.

\begin{table}
\centering
\renewcommand{\arraystretch}{1.25}
\begin{tabular}{|l||c|c|c|c|c|c|}
\hline
      & $k=0$ & $k=1$ & $k=2$ & $k=3$ & $k=4$ & $k>4$\\ 
\hline
$H(x,X)$ & $17$ & $116$ & $102$ & $20$  & $1$ & $0$\\
\hline
$T_8(x,X)$ & $1$ & $132$ & $102$ & $20$ & $1$  & $0$\\
\hline
\end{tabular}
\caption{Number of configurations in $N^*_8(x)$ of a point $x$ which belongs to $X$, 
such that $H(x,X)=k$ or $T_n(x,X)=k$ with $k \in \mathbb{N}$, with $n=8$.}
\label{tab:tab_Hilditch}
\end{table}

\subsection{Yokoi number}

Let $x_0, \ldots, x_7$ be the points of $N^*_8(x)$ 
when traveling through the local neighborhood of $x$ ($N^*_8(x)$), 
starting from a point and returning there, 
and such that 
the first point $x_0$ is a $4$-neigbor of $x$. 
We also define $p_i$, $i \in \{0,\ldots, 7\}$, 
with $p_i=1$ if $x_i \in X$ (for black points) or $p_i=0$ if $x_i \in \overline{X}$ (white points), 
and let $p_8=p_0$.

Let $Y_4(x,X)$ and $Y_8(x,X)$ be the two Yokoi numbers \cite{KoRo89}:
$Y_4(x,X)= \sum_{k \in \{0, 2, 4, 6\}} p_k-(p_k.p_{k+1}.p_{k+2})$, 
and $Y_8(x,X)= \sum_{k \in \{0, 2, 4, 6\}} (1-p_k)-((1-p_k).(1-p_{k+1}).(1-p_{k+2}))$.

In Tab.\ref{tab:tab_Yokoi}, are given the number of configurations 
corresponding to a point $x$ of $X$ such that 
 $Y_n(x,X)=k$ with $k \in \mathbb{N}$;
 we also recall $T_n(x,X)=k$, with $k \in \mathbb{N}$, with $n \in \{4, 8\}$.

There are $116$ configurations such that $Y_n(x,X)=1$, 
we have verified, with a computer, that these configurations 
precisely correspond to $n$-simple points, for $n \in \{4, 8\}$.

\textbf{Proposition$10$:}
Let $X \subset \mathcal{Z}^2$ and $x \in X$, 
$x$ is $4$-simple for $X$ $\Leftrightarrow$ $Y_4(x,X)=1$. 

\textbf{Proposition$11$:}
Let $X \subset \mathcal{Z}^2$ and $x \in X$, 
$x$ is $8$-simple for $X$ $\Leftrightarrow$ $Y_8(x,X)=1$. 

\textbf{Proposition$12$:}
Let $X \subset \mathcal{Z}^2$ and $x \in X$, 
$x$ is $n$-simple for $X$ $\Leftrightarrow$ $Y_n(x,X)=1$, with $n \in \{4, 8\}$. 

We also have, specifically for $n=8$ by using Proposition$8$: 

\textbf{Proposition$13$:}
Let $X \subset \mathcal{Z}^2$. For any point $x$ of $X$, we have $Y_8(x,X)=H(x,X)$.

\begin{table}
\centering
\renewcommand{\arraystretch}{1.25}
\begin{tabular}{|l||c|c|c|c|c|c|}
\hline
      & $k=0$ & $k=1$ & $k=2$ & $k=3$ & $k=4$ & $k>4$\\ 
\hline
$Y_4(x,X)$ & $17$ & $116$ & $102$ & $20$  & $1$ & $0$\\
\hline
$T_4(x,X)$ & $16$ & $117$ & $102$ & $20$ & $1$  & $0$\\
\hline
\hline
      & $k=0$ & $k=1$ & $k=2$ & $k=3$ & $k=4$ & $k>4$\\ 
\hline
$Y_8(x,X)$ & $17$ & $116$ & $102$ & $20$  & $1$ & $0$\\
\hline
$T_8(x,X)$ & $1$ & $132$ & $102$ & $20$ & $1$  & $0$\\
\hline
\end{tabular}
\caption{Number of configurations in $N^*_8(x)$ of a point $x$ which belongs to $X$, 
such that $Y_n(x,X)=k$ or $T_n(x,X)=k$ with $k \in \mathbb{N}$, with $n \in \{4, 8\}$.}
\label{tab:tab_Yokoi}
\end{table}

We also have verified, with a computer, that any configuration in $N^*_8(x)$ is such that $Y_4(x)=T_4(x,X)$ 
except for $1$ configuration: 
this is the configuration corresponding to a $4$-interior point, 
in this case $Y_4(x,X)=0$ and $T_4(x,X)=1$.

\textbf{Proposition$14$:}
Let $X \subset \mathcal{Z}^2$ and $x \in X$. 
$Y_4(x,X)=T_4(x,X)$ except for one single configuration for which $x$ is a $4$-interior point, 
in this case $Y_4(x,X)=0$ and $T_4(x,X)=1$.

With Proposition$9$, we obtain:

\textbf{Proposition$15$:}
Let $X \subset \mathcal{Z}^2$ and $x \in X$. 
$Y_8(x,X)=T_8(x,X)$ except for the $16$ configurations for which $x$ is a $8$-interior point, 
in this case $Y_8(x,X)=0$ and $T_8(x,X)=1$.

\section{Conclusion}\label{sec:conclusion}
 
In this paper, we introduced topological numbers for $2$D binary images $2$D,
we also have characterized simple points with these topological numbers, 
and compare with another existing approaches.

A skeletonization algorithm operates by removing simple points. To be effective, an algorithm must remove multiple simple point configurations simultaneously. 
But, certain configurations of simple points may lead to non-preservation of the topology of the image when these configurations are simultaneously deleted. 
This is the case, for example, if a thinning algorithm deletes the pattern composed of $4$ simple points mutually $8$-adjacent.

Since $116$ configurations lead to a $n$-simple point ($45.31\%$ of possible configurations in $N^*_8(x)$), 
we may say that a parallel thinning of skeletonization 
cannot delete more than $45.31\%$ different configurations of simple points 
during a same iteration while preserving topology.

Our future works consist in studying parallel deletion of specific simple points, 
and we will propose a classification of thinning algorithms according to the (\textit{rate of deletability}), 
which is the number of different local configurations  
that they may delete in a single iteration while preserving topology: 
that will lead to the notion of \textit{more powerful thinning algorithms}.

\bibliographystyle{unsrtnat} 

\end{document}